\documentclass[lettersize,journal]{IEEEtran}
\usepackage{amsmath,amsfonts}
\usepackage{algorithmic}
\usepackage{algorithm}
\usepackage{array}
\usepackage[caption=false,font=normalsize,labelfont=sf,textfont=sf]{subfig}
\usepackage{textcomp}
\usepackage{stfloats}
\usepackage{url}
\usepackage{verbatim}
\usepackage{graphicx}
\usepackage{cite}
\begin{document}

\title{KnobTree: Intelligent Database Parameter Configuration via Explainable \\Reinforcement Learning}

\author{
	\IEEEauthorblockN{
		Jiahan Chen\IEEEauthorrefmark{1},
        Shuhan Qi\IEEEauthorrefmark{1}, 
		Yifan Li\IEEEauthorrefmark{1}, 
		Zeyu Dong\IEEEauthorrefmark{1}
		Mingfeng Ding\IEEEauthorrefmark{2},\\
		Yulin Wu\IEEEauthorrefmark{1},
		Xuan Wang\IEEEauthorrefmark{1}~\IEEEmembership{Senior Member,~IEEE,},
    } 
	\IEEEauthorblockA{\\ \IEEEauthorrefmark{1}Harbin Institute of Technology, Shenzhen\\}
	\IEEEauthorblockA{\IEEEauthorrefmark{2}Tianjin Nanda General Data Technology Co.,Ltd.}

} 

\maketitle

\begin{abstract}
Databases are fundamental to contemporary information systems, yet traditional rule-based configuration methods struggle to manage the complexity of real-world applications with hundreds of tunable parameters. 
Deep reinforcement learning (DRL), which combines perception and decision-making, presents a potential solution for intelligent database configuration tuning. However, due to black-box property of RL-based method, the generated database tuning strategies still face the urgent problem of lack explainability. Besides, the redundant parameters in large scale database always make the strategy learning become unstable. This paper proposes KnobTree, an interpertable framework designed for the optimization of database parameter configuration. In this framework, an interpertable database tuning algorithm based on RL-based differentatial tree is proposed, which building a transparent tree-based model to generate explainable database tuning strategies. To address the problem of large-scale parameters, We also introduce a explainable method for parameter importance assessment, by utilizing Shapley Values to identify parameters that have significant impacts on database performance.  
Experiments conducted on MySQL and Gbase8s databases have verified exceptional transparency and interpretability of the KnobTree model. The good property makes generated strategies can offer practical guidance to algorithm designers and database administrators. Moreover, our approach also slightly outperforms the existing RL-based tuning algorithms in aspects such as throughput, latency, and processing time.
\end{abstract}


\section{INTRODUCTION}
\IEEEPARstart{D}{ATABASES} in the era of big data are facing unprecedented challenges, which have surpassed the capabilities of traditional database technologies. For instance, in the field of image geotagging, efficiently managing and interpreting vast online image databases has become a critical task, yet the sheer size of these databases coupled with the imbalance in data distribution present significant challenges to this endeavor \cite{Trans1Image}. Similarly, within big data infrastructure, bridging the gap between storage performance and application I/O requirements is essential \cite{Trans2Metadata}.

These challenges are indicative of the limitations of traditional database technologies in handling the continuous growth of data volumes and the rapid changes in data characteristics. Firstly, the continuous growth of data volume demands higher processing speeds for individual query tasks. Notably, some recent advancements have focused on optimizing database systems for the efficient handling of temporary data, such as the DiNoDB approach, which streamlines interactive queries on temporary data sets \cite{Trans3Query}. Secondly, the rapid and diverse changes in query loads mean that database configurations and query optimization, based on Database Administrator (DBA) experience, cannot be adjusted in real-time to an optimal runtime state \cite{meng2019survey}. This is primarily due to the fact that database systems have hundreds of adjustable parameters, rendering the system unresponsive to rapid and varied changes. Given these challenges, particularly the difficulties in real-time adjustment of database configurations and handling rapidly changing query loads, traditional database technologies seem inadequate. This situation urgently calls for a more advanced solution, and artificial intelligence, especially Deep Reinforcement Learning (DRL), emerges as a strong candidate due to its potential in handling complex systems and rapidly adapting to changing environments.

Deep reinforcement learning (DRL) has integrated deep learning's perception ability with reinforcement learning's decision-making capability, providing a highly feasible solution for automatic database parameter configuration. Firstly, such methods do not require large amounts of labeled data for training the network, as multiple training samples can be iteratively generated under one load. Secondly, by combining ideas like Markov Decision Processes and gradient descent, the network can quickly fit the target. However, this tuning paradigm presents some challenges: First, existing tuning models only provide a sets of configuration parameters, but the whole decision-making process is a black box, making it impossible to explain the set parameters in the context of business scenarios. When the performance is poor, it is challenging for DBAs to analyze the reasons using past experiences. On the other side, some methods with good interpretability, such as decision trees, may not meet the performance requirements. Secondly, the database systems often have hundreds of parameters for configuration, and considering all of them creates a vast search space, making it difficult to obtain optimal values quickly. Finally, RL-based methods only rely on extensive interactions with the database and neglects utilizing the knowledge of DBA, which is highly inefficient.

This paper addresses the above problems by proposing a interpretable database parameter configuration framework. This framework, called KnobTree, consists of two parts: valueable parameter selection and database tuning decisions. Both of these two parts are designed with good explainability. In the module of important parameter selection, we propose a knowledge-driven and data-driven combined parameter importance evaluation method, using expert knowledge to enhance the database performance prediction under different parameter configurations, and then employing the Sharpley explainable method to screen parameters that significantly influence database performance. In the explainable database tuning decision part, we propose a RL-based tree-structured database tuning algorithm, which allows the model to fully display its decision-making process by constructing a transparent differentiable decision tree model.

The main contributions of KnobTree include three aspects:

\begin{itemize}
    \item We formulate a database parameter configuration model learned by interpretable reinforcement learning. This innovative approach not only ensures optimal database performance but also fully explains the decision-making process, thereby offering coherent explanations for the strategies employed.
    
    \item We introduce a knowledge-driven parameter selection method, which addresses the inefficient issue of excessive parameters in previous approaches. By reducing the action space in reinforcement learning, we are able to accelerate the training process.
   
    \item We engage in comprehensive experimentation across a variety of databases. The empirical evidence underscores the superiority of our proposed method relative to existing techniques, manifesting in enhanced performance and the provision of insightful strategy explanations.
\end{itemize}

\section{RELATED WORK}
Re et al. \cite{re2015machine} first explicitly introduced the concept of integrating machine learning with database systems. Among the technologies discussed, database tuning refers to a category of techniques that optimize performance through system configuration adjustments. Appropriate parameter settings can significantly enhance system performance. There has been substantial research in the area of database tuning \cite{zhu2017bestconfig,li2019qtune,trummer2021case,wang2021demonstrating,van2017automatic,zhang2019end}, with advanced methods capable of learning from historical tasks. Database tuning methods based on learning can be further categorized into those relying on traditional machine learning and those employing deep reinforcement learning.

\subsection{Database Tuning Based on Heuristic Algorithms}

Traditional tuning methods, exemplified by heuristic algorithm-based approaches, are well represented by Zhu et al.'s system BestConfig \cite{zhu2017bestconfig}. The core idea of BestConfig is to utilize heuristic algorithms to search the parameter space. The algorithm initially discretizes the parameter space and then employs a bounded recursive search algorithm to randomly select a set of samples for testing in each iteration. It identifies the best-performing point and uses this point as the center to define the sampling region for the next iteration. This process continues until no better performing point can be found. However, this method explores the parameter space through sampling, making it challenging to find better solutions in vast parameter spaces. Moreover, it fails to fully utilize historical data. Each tuning process starts from scratch, without the ability to optimize the model based on previous training, leading to significant resource and time wastage.

\subsection{Database Tuning Based on Machine Learning}

Database tuning methods based on traditional machine learning conceptualize the tuning task as a regression problem. Dana et al. proposed the OtterTune system \cite{van2017automatic,zhang2018demonstration}, which employs Gaussian Process Regression \cite{seeger2004gaussian} to identify optimal parameters. Gaussian Process Regression treats the relationship between configuration parameters and workload as a multivariate Gaussian joint distribution, balancing exploration and exploitation. OtterTune begins by using factor analysis \cite{shrestha2021factor} to filter irrelevant features. It then employs the K-means clustering method \cite{ahmed2020k} to select a few features most closely related to the parameters, serving as inputs to the model. At this stage, OtterTune runs the Gaussian Process model under the current workload, generating a set of random variable values associated with the inputs and following a Gaussian distribution. These values are recommended as the optimal parameters for the database system under the current load, while simultaneously optimizing the tuning model.

To address the issue of vast parameter spaces in DBMS, Kanellis et al. \cite{kanellis2020too} introduced a method to reduce the number of parameters needing adjustment through pre-selection, employing the Classification and Regression Tree (CART) \cite{loh2011classification} technique to quantify the importance of each parameter to be tuned.
Cereda et al. developed CGPTuner \cite{cereda2021cgptuner}, drawing inspiration from hyperparameter optimization in machine learning. This approach utilizes Bayesian optimization to search for the optimal configuration in the current context.
Nguyen et al. \cite{nguyen2018towards} adopted Latin Hypercube Sampling (LHS) \cite{cheng2000latin} to minimize the search space. They trained a simple machine learning model to predict the performance of a specified application under recommended configurations. Finally, they used a recursive random search method to identify the best parameter configurations. This multi-faceted approach reflects the diverse strategies in the field to efficiently navigate the expansive parameter spaces of DBMS.

While traditional machine learning systems exhibit strong generalization capabilities and perform well in various database environments, these methods still have significant limitations. First, these methods typically employ a pipeline architecture, where the optimal solutions obtained at each stage may not necessarily be globally optimal. Moreover, the models at different stages might not effectively complement each other.
Second, these methods require a substantial amount of high-quality samples for training the models, but such samples are often difficult to obtain. Additionally, relying solely on Gaussian Process Regression models proves inadequate for representing the problem of database tuning, which involves high-dimensional continuous spaces. This limitation stems from the model's inherent complexity in adequately capturing the nuances of such a vast parameter space.

\subsection{Database Tuning Based on Deep Reinforcement Learning}

Database tuning methods based on deep reinforcement learning involve an iterative process of trial and error through interactions between an agent and the database environment. This approach continuously optimizes the strategy by which the agent selects parameters. One of the key advantages of this method is that it does not require a large amount of labeled data. Additionally, it leverages exploration and exploitation mechanisms, striking a balance between venturing into unknown parameter spaces and capitalizing on existing knowledge. In recent years, most research on automatic database tuning has predominantly focused on this approach.

Zhang and colleagues developed an end-to-end tuning system, CDB-Tune \cite{zhang2019end}, specifically designed for online tuning in cloud database scenarios. After collecting workload data, CDB-Tune utilizes a deep reinforcement learning model to recommend database parameters and record performance metrics. It employs an offline-trained model for online adjustments, concurrently updating the deep reinforcement learning model and the memory pool. This approach significantly enhances the efficiency of tuning.
Gur and team \cite{gur2021adaptive} proposed a multi-model online tuning method, which initially trains multiple models for different workloads. From these, the model most likely to increase rewards is selected for transfer learning. This new model is then used to generate configurations, replacing the old model. This strategy of training multiple foundational models addresses the issue of tuning performance instability due to workload variations.

\subsection{Explainable Machine Learning}

In the field of explainability, Shapley values \cite{lundberg2017unified,cohen2005feature} are a key method for quantifying parameter importance. They provide a mathematically sound and equitable approach to interpret machine learning decisions, enhancing transparency and trust in model predictions.

To acquire imprecise knowledge in ambiguous environments, some fuzzy decision trees have been developed. The fuzzy decision tree proposed by Suarez et al. \cite{suarez1999globally} is an extension of the traditional decision tree, enabling it to handle uncertainties and thus possessing stronger classification capabilities and robustness. In 2017, Hinton introduced a method to map deep neural networks onto differentiable decision trees, providing a vivid and rational interpretation of neural networks \cite{frosst2017distilling}. In 2019, Andrew and others introduced Prolonets \cite{silva2021encoding}, which allows for the initialization of neural network architecture using domain knowledge and uses policy gradient updates for parameters, making full use of expert domain knowledge.

Despite the recent advances in deep reinforcement learning, there remain pressing issues to be resolved. One issue is the excessive action space due to redundant information in database parameters, which the end-to-end deep reinforcement learning methods fail to preprocess, leading to longer training times. Another critical concern is the lack of interpretability in existing reinforcement learning methods, a significant shortcoming in database configuration. To address these challenges, this study proposes an explainable reinforcement learning-based approach for database tuning. By integrating a knowledge-driven performance prediction model with Shapley additive explanations \cite{cohen2005feature}, the method not only effectively tunes databases but also provides logical explanations for its tuning strategies.

\section{PRELIMINARIES}

\subsection{Shapley Value}

The Shapley value \cite{lundberg2017unified} is one of the cooperative game theory solutions, which assigns a unique distribution of a total surplus generated by the coalition of all players among the players. It represents the average marginal contribution of a player by collaborating with others. The Shapley value of player \( j \) for game \( v \) can be defined by:

\begin{equation}
\phi_j(v) = \sum_{S \subseteq \{1, \ldots, p\}} \frac{|S|!(p - |S| - 1)!}{p!} \left( v(S \cup \{j\}) - v(S) \right)
\label{eq_shap}
\end{equation}

where \( S \) is a subset of players excluding player \( j \), \( p \) is the total number of players, and \( v(S) \) is the value that the coalition \( S \) could obtain by itself. The Shapley value of player \( j \) is the weighted sum of marginal contributions across all coalitions.

In the domain of explainable machine learning, SHAP (Shapley Additive Explanations) values represent a method for explaining the output of machine learning models\cite{verhaeghe2022powershap}. They are based on the concept of Shapley values from cooperative game theory. The SHAP value is a model-agnostic measure and is defined as follows:

\begin{equation}
g(z') = \phi_0 + \sum_{j=1}^{M} \phi_j z_j'
\end{equation}

Here, \( g \) is the model function, \( z' \in \{0,1\}^M \) is a binary vector indicating the presence or absence of features, \( M \) is the total number of features, and \( \phi_0 \) is the model output for the absence of all features.

KernelSHAP is an algorithm for explainable machine learning that approximates SHAP values efficiently. The procedure is as follows:

\begin{enumerate}
    \item Generate feature permutations;
    \item For each permutation \( z' \), calculate the corresponding SHAP value;
    \item Assign a weight to each \( z' \);
    \item Sum the weighted SHAP values to obtain the final explanation of the model.
\end{enumerate}

The weight assigned to each permutation is given by:

\begin{equation}
    \pi_{x}(z') = \frac{M-1}{\binom{M}{|z'|} |z'| (M - |z'|)}
\end{equation}

where \( M \) is the total number of features, and \( |z'| \) is the number of non-zero features in the permutation. The final SHAP value is calculated as:

\begin{equation}
    L(f,g,\pi_x) = \sum_{z'\in Z} [f(h_x(z'_k)) - g(z')]^2 \pi_x(z')
    \label{eq1}
\end{equation}

The concept of using SHAP for feature selection is based on the idea that features corresponding to larger absolute Shapley values are more important. To implement this, one computes the mean absolute Shapley value for each feature across the entire dataset. Then, by sorting the features in descending order based on these means, one can select the most important features.

\subsection{Reinforcement Learning}

Reinforcement learning for database tuning optimizes the strategy of parameter selection through trial and error interactions between an agent and the database environment. In the framework of reinforcement learning, the agent interacts with the environment, takes an action, changes the state of the environment, receives feedback from the environment, i.e., the reward, and adjusts its strategy based on the reward received.

A fundamental concept in reinforcement learning is the tuple \( (S, A, T, R, \gamma) \), which represents the state space, action space, transition function, reward function, and discount factor, respectively. The reward received by the agent at each time step is defined by:

\begin{equation}
    G_t = r_{t+1} + \gamma r_{t+2} + \gamma^2 r_{t+3} + \ldots + \gamma^{T-t-1} r_T
\end{equation}

where \( T \) is the time at the end of the episode, representing the cumulative reward from the current state to the end of the episode. Besides, the action-value function \( Q_\pi(s, a) \) represents the expected return of taking action \( a \) in state \( s \) under policy \( \pi \):

\begin{equation}
    Q_\pi(s, a) = \mathbb{E}_\pi [ G_t | S_t = s, A_t = a ]
\end{equation}

Reinforcement learning is divided into two approaches: value-based learning and policy-based learning. Value-based methods select the action that maximizes the value function at each decision point, with the goal of learning the optimal value function. Policy-based methods calculate the probability distribution of actions in each state according to a policy function, with the objective of learning the policy function itself.

Konda proposed a reinforcement learning algorithm widely used for continuous action control problems: Actor-Critic \cite{konda1999actor}. This algorithm combines the advantages of both value-based and policy-based learning methods, consisting of two modules: the actor and the critic. The actor module corresponds to the policy function and is responsible for learning high-return strategies, while the critic module corresponds to the value function and is used to estimate the value of the current policy. Each time the actor module outputs an action, the critic module evaluates the return of that action to assess its quality. By jointly training the actor and critic modules, the Actor-Critic algorithm can achieve good performance on continuous action control problems.

Deep Deterministic Policy Gradient (DDPG) \cite{lillicrap2015continuous} is a deep reinforcement learning algorithm designed for continuous action control problems, representing an extension of the actor-critic architecture. The optimization objective of the DDPG algorithm is to update the neural network parameters by minimizing the loss functions of the value function and the policy function, while utilizing target networks to reduce jitter during the update process. The interpretable intelligent tuning method presented in this paper is based on an improvement of the DDPG algorithm. This method enhances the algorithm's interpretability and transparency by incorporating an interpretable actor module into the foundation provided by the DDPG algorithm.

\section{KnobTree Architecture}

Database configuration parameters may contain redundant information. Traditional reinforcement learning tuning methods have failed to preprocess these configuration parameters, either treating all parameters as tunable or arbitrarily designating only a few for tuning. The former approach creates an excessively large action space in reinforcement learning, hindering training convergence, while the latter heavily depends on prior knowledge and struggles to adapt across different database system environments. In response to these challenges, KnobTree introduces a method for parameter selection. This approach begins by constructing a knowledge-driven database performance prediction model. Subsequently, it employs Shapley Additive Explanations to calculate each configuration parameter's contribution to performance, selecting those with the most significant impact as the tunable parameters.

Existing database tuning methods have only focused on the performance of intelligent tuning, neglecting the importance of explainable parameter configuration. However, an explainable intelligent tuning scheme is instructive for both model designers and database administrators, and is especially vital in a fault-intolerant production environment. Therefore, KnobTree contains a tuning method based on explainable reinforcement learning. We propose an Actor-Critic reinforcement learning architecture based on differentiable decision trees, employing a transparent tree model for decision-making, thus generating explainable parameter configuration strategies.

\begin{figure*}[t]
\centering
\includegraphics[width=4.5in]{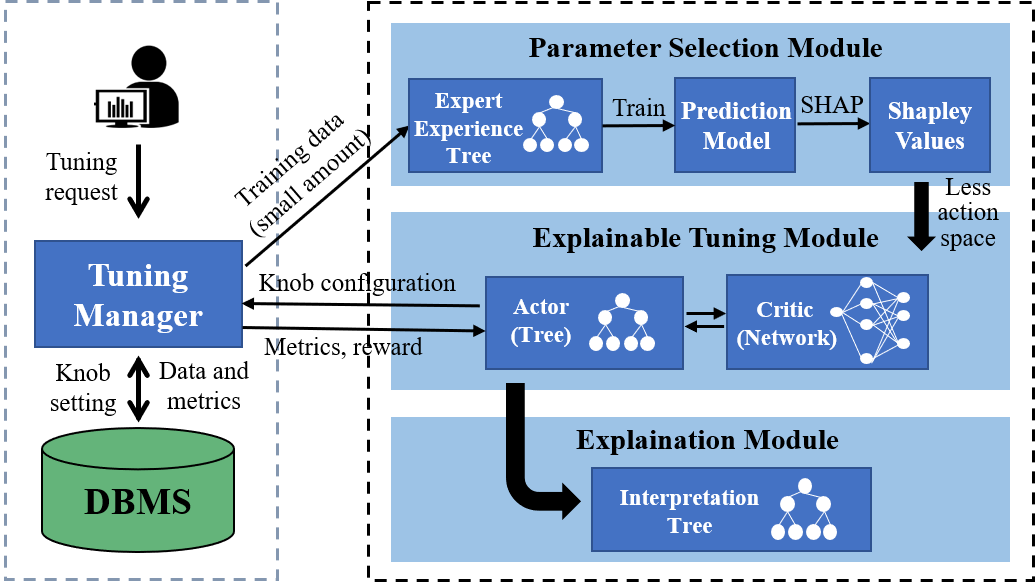} 
\caption{The overall model of KnobTree. The parameter selection module requires only a small amount of data to construct a predictive model, filtering parameters based on Shapley values. The tuning module consists of a tree-shaped actor network and a standard critic network. The explanation module transforms the tuning module's actor network into an explanation tree.}
\label{fig1}
\end{figure*}

The overall architecture of the interpretable database tuning method (KnobTree) designed in this paper is divided into three modules. First, an importance analysis of the database parameters is conducted to select the parameters that have the most significant impact on performance for model training. Next, an interpretable tuning model is used for parameter recommendation. Finally, an explanation tree is generated to interpret the strategy, as shown in Fig.~\ref{fig1}.

\section{Explainable Tuning Techniques}

Core contributions of KnobTree encompass two main parts: knowledge-driven configuration parameter selection and optimization of parameters based on explainable reinforcement learning.

\subsection{Knowledge-driven parameter selection}

\begin{figure*}
	\centering
	\subfloat[Expert prior knowledge]{ \includegraphics[width=2in]{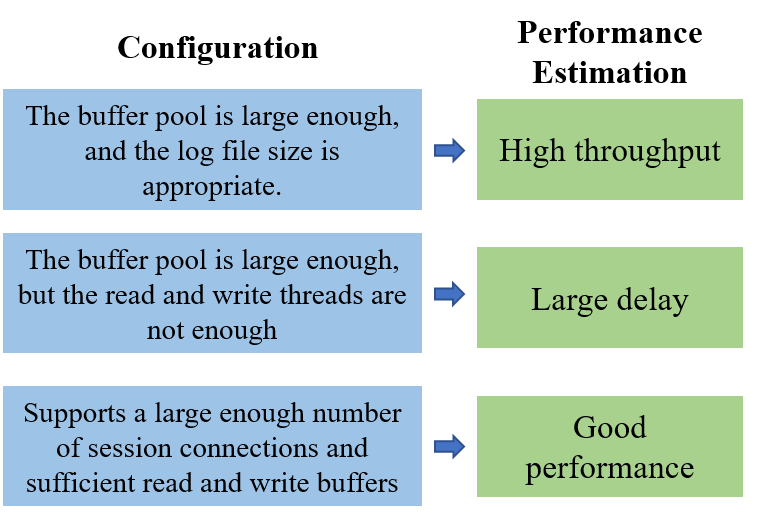} \label{Fig2(a)}}\hspace{0.5in}
	\subfloat[Prediction tree]{ \includegraphics[width=2in]{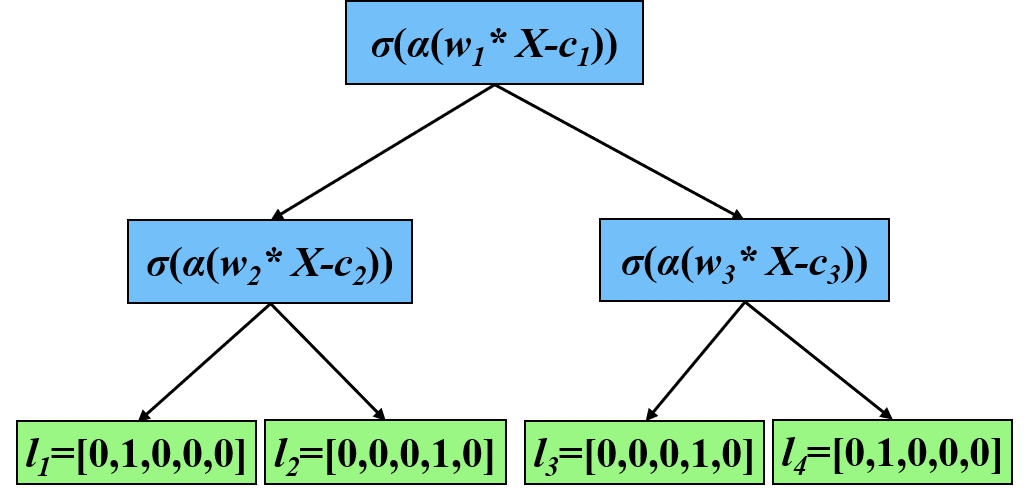} \label{Fig2(b)}}
	
	\quad
 
	\subfloat[Training on dataset]{ \includegraphics[width=2in]{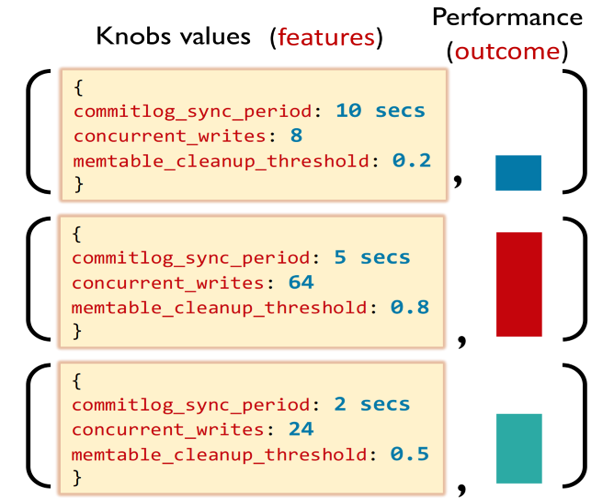} \label{Fig2(c)}}\hspace{0.5in}
    \subfloat[Importance analysis]{ \includegraphics[width=2in]{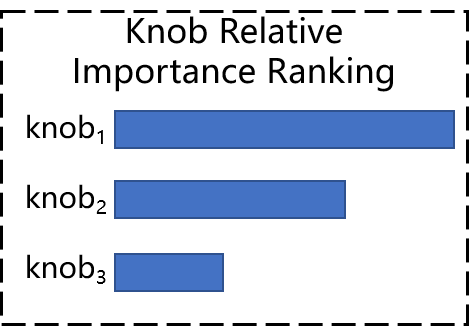} \label{Fig2(d)}}
 
	\caption{The workflow of parameter selection. (a) Gathering expert knowledge; (b) Transforming this knowledge into a prediction tree; (c) Constructing a dataset and training the prediction model; (d) Analyzing parameter importance using Shapley values.}
	\label{Fig2}
\end{figure*}

The whole method of parameter selection by KnobTree is depicted in Fig.~\ref{Fig2}. Due to the scarcity of high-quality data for the database parameter tuning task, we designed a knowledge-driven performance prediction model, using the expert knowledge shown in Fig.~\ref{Fig2(a)}, constructing a decision tree, and then converting the nodes in the tree into neural network parameters to facilitate training. We utilized Latin Hypercube Sampling to gather data and trained decision trees. Finally, we employed the Shapley Additive Explanations method to calculate the contribution of each configuration parameter to performance, thereby identifying a set of the most critical parameters to serve as the action space for the reinforcement learning model. Our developed model was designed with transferability in mind. By employing a knowledge-driven approach that combines expert understanding and domain-specific insights, the model is enabled to identify fundamental patterns that may be applicable across various scenarios.

\subsubsection{Prediction Model}

\begin{figure}[t]
\centering
\includegraphics[width=0.95\columnwidth]{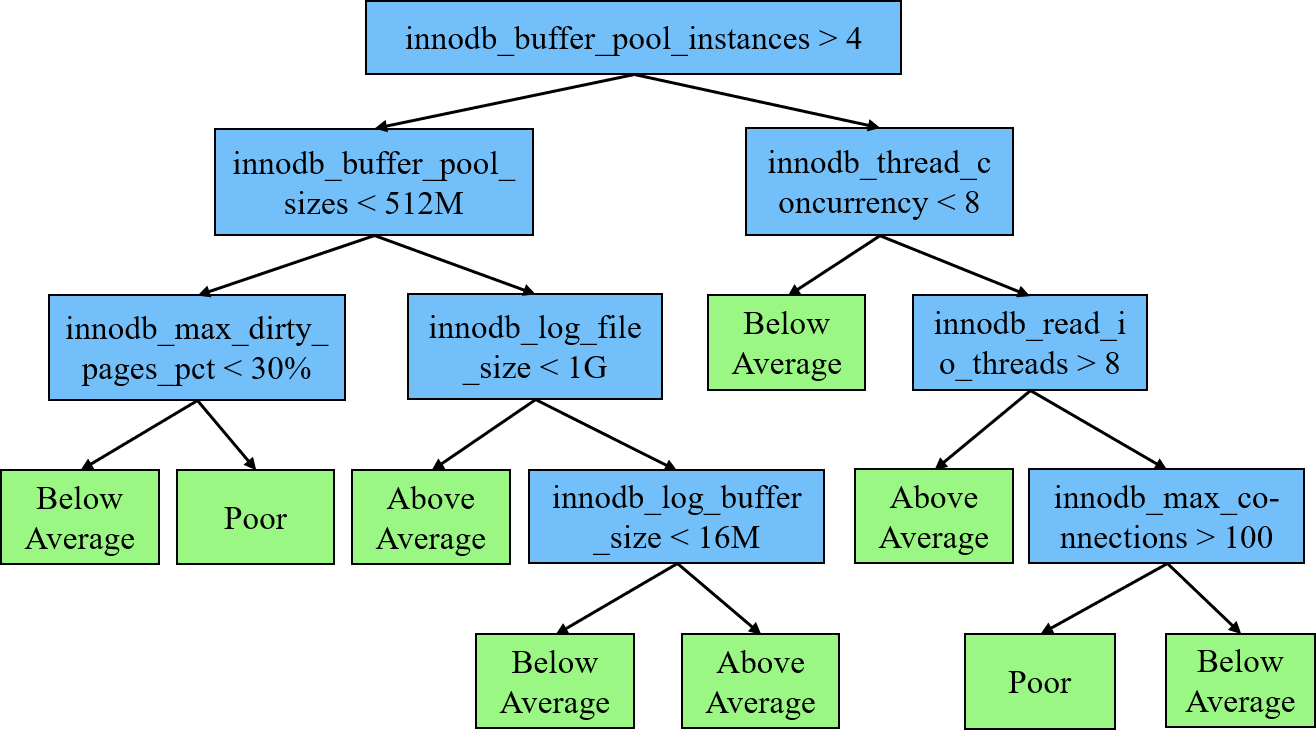} 
\caption{Structure of expert experience tree. Internal nodes (blue) correspond to rules, while leaf nodes (green) correspond to predicted performance.}
\label{fig3}
\end{figure}

The knowledge-driven prediction model can utilize experiential knowledge to initialize the parameters of the model, thereby reducing training time. We convert expert experience into a hierarchical set of rules, represented as a tree-like structure. Figure~\ref{fig3} illustrates an example of the expert experience tree.

To facilitate the evaluation of database performance, we categorize it into several levels: "Poor," "Below Average," "Average," "Above Average," and "Excellent." The model predicts the performance level based on the input configuration parameters. The internal nodes of the experience tree correspond to the rules for performance evaluation, each composed of a configuration parameter and a threshold. The decision to move to a particular branch is determined by comparing these two elements, and the leaf node reached at the end corresponds to the predicted performance level.

In order to be able to carry out effective training, we convert the experience tree into a tree-shaped neural network. The experience in the tree is transformed into a series of rules composed of weight $w$ and comparison threshold $c$. The weight $w$ indicates which configuration parameter to judge according to, and the threshold $c$ provides the criterion for judgment. For internal node $D_i$, each rule is expressed as: 
\begin{equation}
D_i = \sigma [\alpha (w_i X - c_i)]
\label{eq2}
\end{equation}
Here, $X$ is the input configuration parameter vector, $\alpha$ is a hyperparameter used to limit the credibility of internal nodes. Thus, each layer of internal nodes in the tree corresponds to a hidden layer in the neural network. The leaf nodes represent the probability of each performance level, initialized as corresponding one-hot vectors. For example, [0, 0, 0, 1, 0] indicates the performance level "Above Average". The converted tree structure is shown in Figure 2(b).

The resulting neural network prediction model can be expressed as:
\begin{equation}
P = \sum_{j\in Leaf} \left( W_j \prod_{i\to j} D_i \right) 
\label{eq3}
\end{equation}
Here, $P$ is the database performance, and $W_j$ is the weight of the leaf node. Each leaf node corresponds to a path, and the possibility of reaching the leaf node can be obtained by multiplying the possibility of the internal nodes on the corresponding path. The possibility of each leaf node is multiplied by the corresponding weight to get the final performance.

\subsubsection{Training}

At the current stage, although a parameter prediction model has been established, it heavily relies on expert knowledge and requires a substantial amount of data for training. Given the current lack of publicly available datasets associating database parameters with performance, it becomes necessary to construct a dedicated dataset for the task of database tuning.

The dataset for the database tuning task consists of tuples containing configuration parameters and system throughput. The construction process of the dataset is as follows:

In our study, we utilized a Latin hypercube sampling technique, as discussed in the OtterTune paper \cite{van2017automatic}, to efficiently sample the multi-dimensional space of database configuration parameters. This method allowed us to obtain a comprehensive dataset by deploying these parameters in MySQL and Gbase8s databases and measuring performance using evaluation tools. Preprocessing steps like the removal of enumeration types and normalization were applied to prepare the data for our prediction model.

The empirical tree prediction model was then trained with this data, predicting average performance for each configuration. We employed cross-entropy loss and the Adam optimizer to enhance the model's accuracy and optimization process, ensuring effective performance prediction across the database configurations.

\subsubsection{Parameter Importance Analysis}

We have successfully built a model that can predict performance based on database configuration parameters, allowing us to obtain any data sample. For each data sample, we consider the database performance as the sum of the contribution values of each configuration parameter, and use the Shapley additive explanation method to determine the contribution value of each parameter. This helps us reduce the dimensionality of the action space for the next stage of the model.

To calculate the Shapley value of a parameter, we first need to select several parameters from the sample's configuration parameter set to form a coalition, represented by an indicator vector $Z$ composed of 0's and 1's, where, for example, Z = {1, 0, 1, 0} represents that the first and third configuration parameters are included in the alliance. The values of the configuration parameters not selected in the coalition are replaced by the values of those parameters randomly sampled from the dataset. 

In this paper, KernelSHAP is used to calculate weights, followed by optimizing the loss function in (\ref{eq1}) to train a linear model, which serves as an explanation for the contribution of parameters in the predictive model. Configuration parameters with higher Shapley values are considered more important for system performance. Since global importance needs to be taken into account, the absolute values of the Shapley values for each configuration parameter are averaged across the entire dataset:

\begin{equation}
I_j = \frac{1}{n} \sum_{i=1}^n \left| \Phi_j^{(i)} \right|
\end{equation}

\begin{figure}[t]
\centering
\includegraphics[width=0.95\columnwidth]{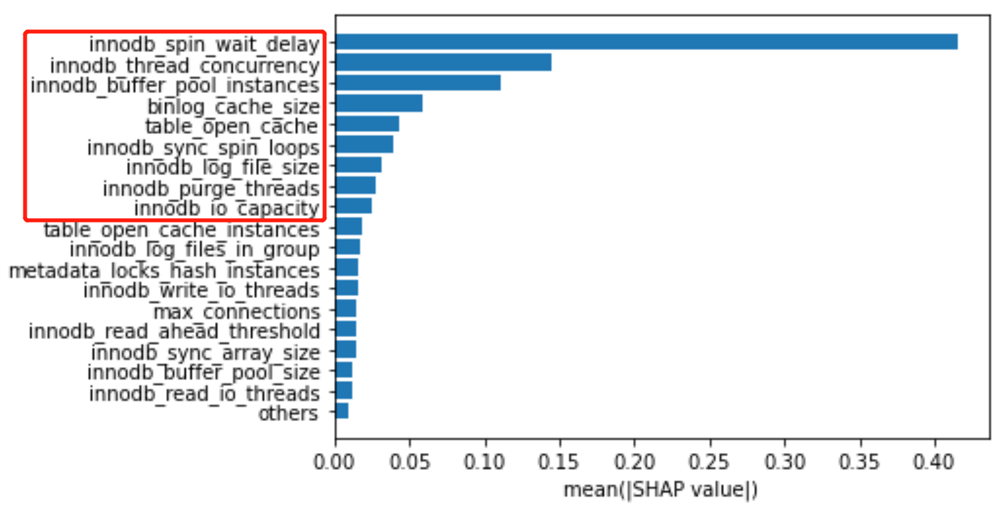} 
\caption{Parameter screening using SHAP values on MySQL. The red box highlights the nine most important parameters selected through the analysis, representing key influences on database performance.}
\label{fig4}
\end{figure}

Finally, the configuration parameters are sorted in descending order of importance, allowing us to select the most important ones. The parameters finally screened out in MySQL are shown in Figure~\ref{fig4}, for instance.

\subsection{Parameter Tuning Based on Explainable RL}

The elements of reinforcement learning can correspond to the elements of the database tuning task: the agent represents the tuning model, the environment represents the database system, actions correspond to changes in certain parameters, states correspond to the internal performance indicators of the database, and rewards correspond to performance changes after deploying new configurations.

Existing reinforcement learning-based tuning methods, such as CDBTune \cite{zhang2019end}, have been able to achieve good performance. However, these methods can only generate a set of policies and do not provide explanations for these policies. This lack of interpretability makes it challenging for database administrators to maintain configurations when issues arise.

Based on the DDPG architecture, we modify the actor module responsible for generating policies to a differentiable decision tree. On one hand, the actor tree model offers good transparency and interpretability, for the process of policy generation can be explained by tracing the path from leaf nodes to the root node. On the other hand, the actor module differs from traditional decision trees that can only make predictions. Instead, it can generate policies starting from the root node and participate in backpropagation training, which significantly enhances the practical performance of the decision tree.

\begin{figure}[t]
\centering
\includegraphics[width=0.90\columnwidth]{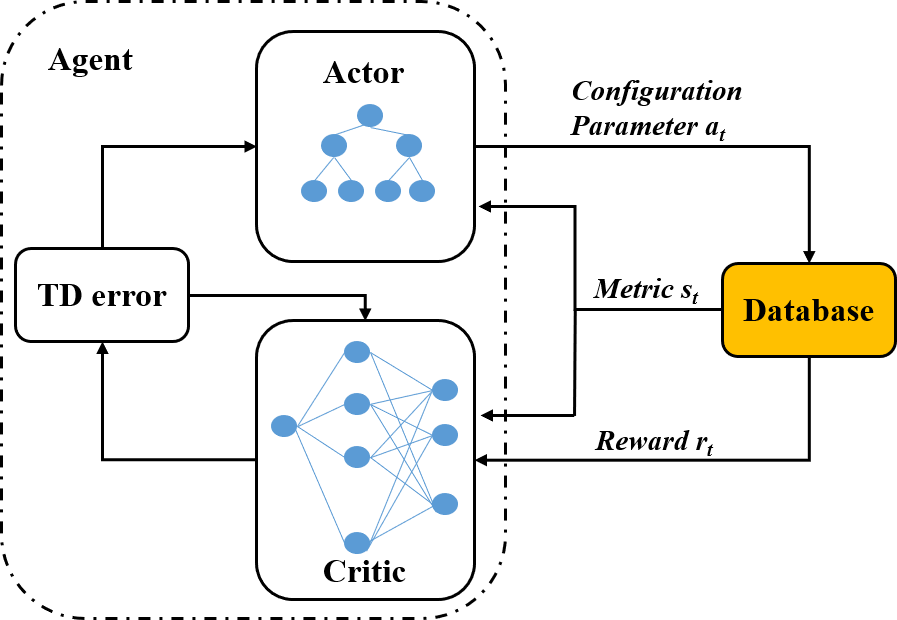} 
\caption{Explainable tuning model structure. The overall architecture aligns with other RL-based tuning methods, but the actor is implemented by a differentiable decision tree.}
\label{fig5}
\end{figure}

The structure of the entire tuning model is shown in Figure~\ref{fig5}. The actor generates a set of policies, namely configuration parameters, which when applied to the database cause changes in the indicators. The critic also gives corresponding scores, and the model calculates new rewards to train these two modules.

\subsubsection{Explainable Actor}

The input of the actor module is the observed internal indicators of the database, and the output is the configuration parameters. The leaf nodes in the tree serve as inputs, internal nodes replace the traditional decision tree's boolean expressions with linear combinations of input features, and the root node serves as the output. Input data propagates from the leaf nodes upwards, allowing for the updating of node parameters through backpropagation during reinforcement learning training.

In computing the internal node weights, the Gumbel-Softmax method \cite{jang2016categorical} is introduced to control the weight distribution, avoiding overly uniform weights and enhancing the model's flexibility and expressiveness. The algorithmic process of the parameter generation module is as shown in Algorithm \ref{alg1}.

\begin{algorithm}[tb]
\caption{Explainable Parameter Generation Algorithm}
\label{alg1}
\textbf{Input}: Database internal performance vector $S$\\
\textbf{Parameter}: Internal node parameter$(\alpha, \mathbf{W}, \mathbf{C})$,tree height $h$\\
\textbf{Output}: Configuration vector $\mathbf{iter}$
\begin{algorithmic}[1]
\STATE Initialize leaf node $\mathbf{leaf}$, with dimension $2^{h-1}$.
\STATE Initialize iteration vector: $\mathbf{iter = leaf}$.
\FOR{$i = h-1$ to $0$}
    \FOR{$j = 1$ to $2^{h-1}$}
        \STATE Compute $\mathbf{\sigma} = {sigmoid}(\alpha_{ij} \cdot \mathbf{(W_{ij}S - C_{ij})})$.
        \STATE Compute $\mathbf{iter} = \sigma \cdot \mathbf{iter}_{2j} + (1-\sigma) \cdot \mathbf{iter}_{2j+1}$.
        \STATE Update $\mathbf{iter}$.
    \ENDFOR
\ENDFOR
\STATE Output $\mathbf{iter}$.
\end{algorithmic}
\textit{Note:} In this process, the dimension of $\mathbf{iter}$ will be halved in every iteration, until it reaches the root.
\end{algorithm}

\subsubsection{Training}

We use the same reward function as CDBTune \cite{zhang2019end}. The reward function should consider not only the changes in database performance compared to the last parameter tuning, but also the changes in performance compared to the initial configuration. Specifically, the changes in throughput and delay in these two time periods are calculated, recorded as $\delta_{t\to t-1}$ and $\delta_{t\to 0}$, separately.

If the performance change rate for either throughput or delay compared to the initial moment is greater than 0, the tuning trend is correct and the reward is positive; if less than 0, the reward is negative:
\begin{equation}
r = \left\{
\begin{aligned}
    ((1+\delta_{t\to 0})^2 - 1)|1+\delta_{t\to t-1}| \quad \delta_{t\to 0} > 0 \\
    -((1-\delta_{t\to 0})^2 - 1)|1-\delta_{t\to t-1}| \quad \delta_{t\to 0} \leq 0 \\
\end{aligned}
\right.
\label{eq5}
\end{equation}

Now we can calculate rewards $r_T$ and $r_L$ corresponding to the throughput and delay of the database system separately. Adding the weight $C_L+C_T=1$ to deal with the different requirements of throughput and delay in actual scenarios, the final reward function $R$ is as follows:
\begin{equation}
R = C_T * r_T + C_L * r_L
\label{eq6}
\end{equation}

The critic module is trained using Temporal Difference (TD) error, and then the actor module is trained through backpropagation (policy gradient), based on the critic module. This training approach is the same as that of DDPG \cite{lillicrap2015continuous}.

\subsubsection{Interpretation Tree Generation}

\begin{figure}[t]
\centering
\includegraphics[width=0.95\columnwidth]{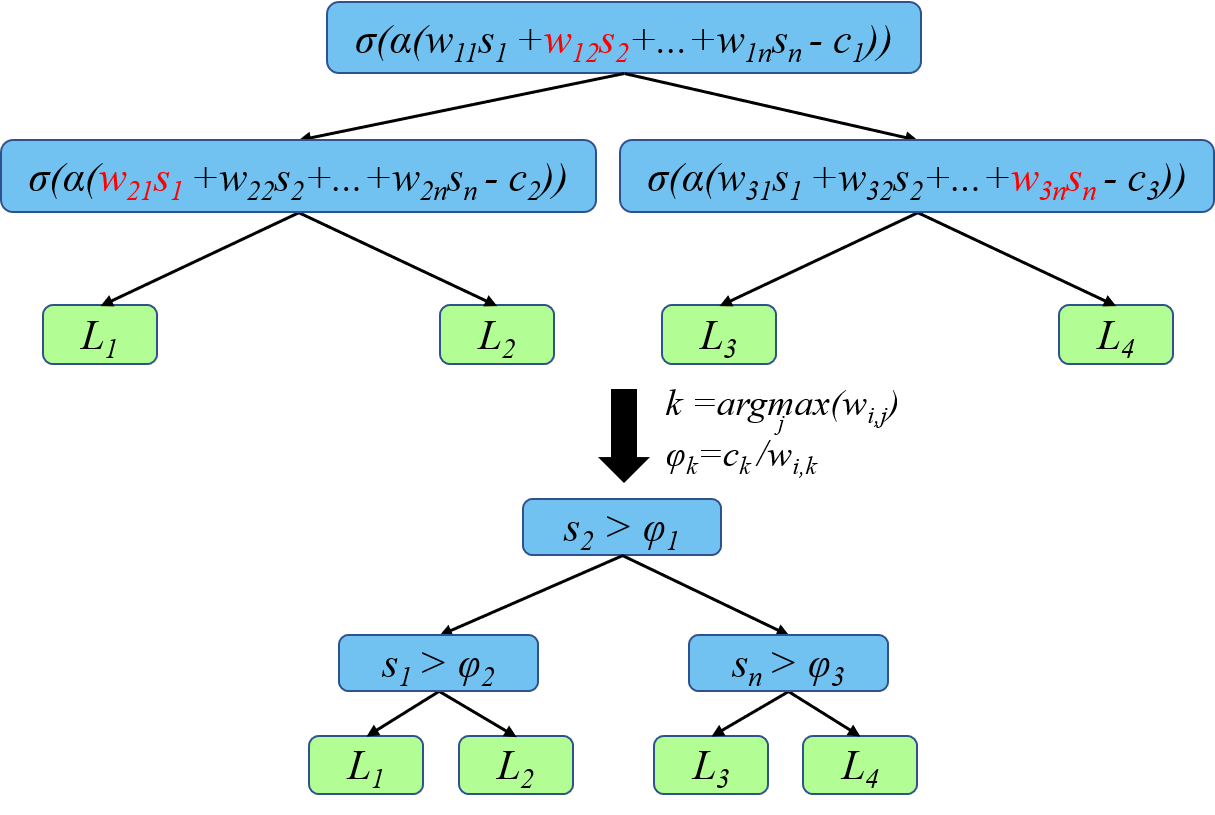} 
\caption{Generation of policy interpretation tree. Each internal node is converted into a traditional decision tree node by selecting the most important feature and normalizing threshold values. }
\label{fig6}
\end{figure}

After the training is completed, the tree-structured actor can be discretized to extract the embedded useful information, thereby interpreting the parameter configuration.
Specifically, at this point, the weight and threshold of each feature in the inner nodes of the tree have been determined. The most crucial feature in that node can be obtained through $argmax$. Subsequently, the thresholds in the internal nodes can be normalized by dividing by the weight of the selected feature, transforming each internal node into a traditional decision tree node. This transformation allows for the interpretation of the node as comparing a single feature against a threshold. In the discretized tree, the path from the root node to the leaf node elucidates the decision-making process of the actor. Figure~\ref{fig6} illustrates an example of node conversion and the generation of an explanatory tree for tuning strategy.

\section{Experimental Evaluations}

Our experimental equipment is based on the Ubuntu18.04 system, with Intel(R) Core(TM) i9-9900X CPUs and NVIDIA GeForce RTX 2080Ti GPUs. This paper conducts model evaluation on two databases, MySQL and Gbase8s. For the former, the sysbench pressure testing tool is used to test the performance, and for the latter, the benchmarkSQL pressure testing tool is used. 

The most commonly used metrics for measuring database performance are throughput and latency. Throughput represents the number of transactions processed by the database per unit of time, measured based on the number of successful submissions. Latency, on the other hand, refers to the response time of the database to client requests, measuring the time taken by the database service to process requests. A widely adopted metric for latency is the 95th percentile latency time, which signifies the latency duration for the top-performing 95\% of requests. For example, if the 95th percentile latency time is 2 seconds, it means that 95\% of requests can be processed within 2 seconds. In this paper, throughput and the 95th percentile latency time are used as two metrics to evaluate database performance from the perspectives of processing capacity and processing speed, respectively.

\subsection{Tuning Effects Comparison}

Experiments in this paper primarily focus on OLTP benchmark tests, which support various workload scenarios, including read-only, write-only, and read-write scenarios. The composition of transactions in these three scenarios is as follows:
\begin{itemize}
    \item In the read-only scenario (RO), one transaction consists of 14 read SQL operations (10 primary key point queries and 4 range queries).
    \item In the write-only scenario (WO), one transaction consists of 4 write SQL operations (2 \texttt{UPDATE} operations, 1 \texttt{DELETE} operation, and 1 \texttt{INSERT} operation).
    \item In the read-write mixed scenario (RW), one transaction consists of 18 read-write SQL operations.
\end{itemize}

This paper evaluates the effectiveness of the following well-known database tuning approaches after they have converged:

\begin{itemize}
  \item \textbf{KnobTree} (proposed in this paper): An explainable parameter tuning method using parameter selection, where the input of action consists of the selected parameters.
  \item \textbf{CDBTune}: An end-to-end automatic parameter tuning method based on reinforcement learning, trained offline and recommends parameters online.
  \item \textbf{OtterTune}: An automatic parameter tuning method based on large-scale machine learning, which trains a performance prediction model and selects the best configuration.
  \item \textbf{MySQLTuner}: A MySQL configuration parameter recommendation tool written in Perl (not applicable in the Gbase8s experiment).
  \item \textbf{DBA}: Database administrator tuning parameters based on experience.
  \item \textbf{Default}: System performance under default configurations.
\end{itemize}

\begin{figure*}
	\centering
	\includegraphics[width=2.5in]{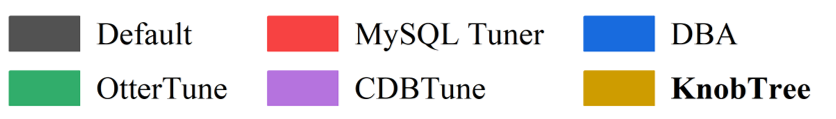} \hspace{0.5in}
	\quad
 
	\subfloat[RO(Throughput)]{ \includegraphics[width=2in]{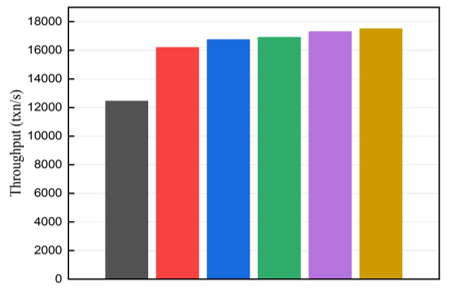} \label{Fig7(a)}}
	\subfloat[WO(Throughput)]{ \includegraphics[width=2in]{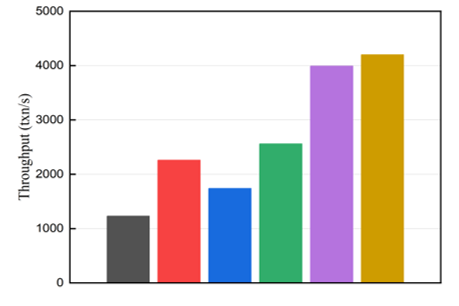} \label{Fig7(b)}}
	\subfloat[RW(Throughput)]{ \includegraphics[width=2in]{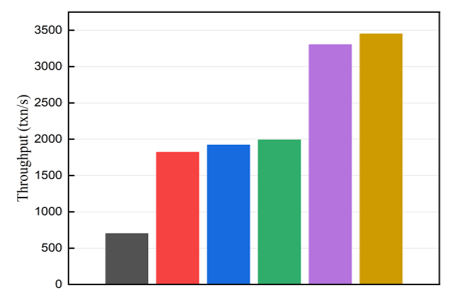} \label{Fig7(c)}}
	
	\quad
 
	\subfloat[RO(Delay)]{ \includegraphics[width=2in]{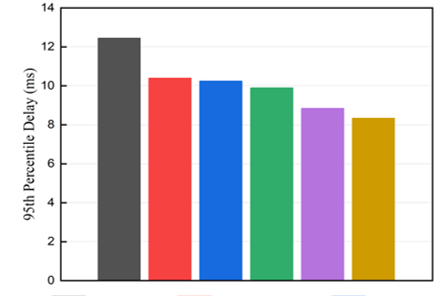} \label{Fig7(d)}}
    \subfloat[WO(Delay)]{ \includegraphics[width=2in]{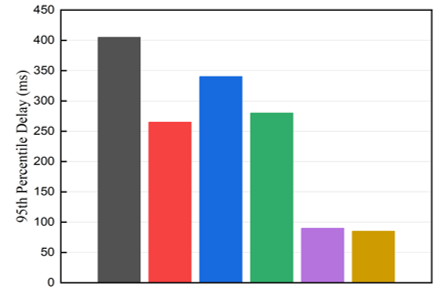} \label{Fig7(e)}}
    \subfloat[RW(Delay)]{ \includegraphics[width=2in]{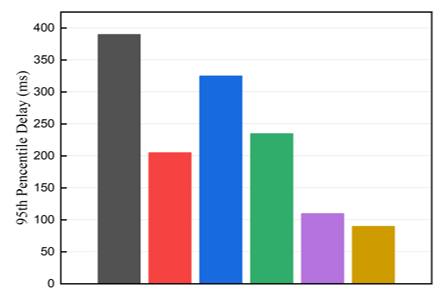} \label{Fig7(f)}}
 
	\caption{Comparison of tuning results on MySQL.}
	\label{Fig7}
\end{figure*}

It is noteworthy that CDBTune can be regarded as an ablation experiment of the differentiable decision tree in KnobTree, since both utilize the DDPG architecture, with the only difference being in the Actor component. Apply all the above methods to the read-only, write-only, and read-write scenarios of MySQL respectively. The tuning effect is shown in Figure~\ref{Fig7}. 

Compared with the default configuration parameters of the database, the above several intelligent parameter adjustment methods have obvious performance improvements. Compared with OtterTune, the method in this paper has achieved obvious advantages, because OtterTune is based on large-scale machine learning and requires a large amount of high-quality training data. The method in this paper is based on reinforcement learning, which can continuously explore trial and error in the environment, making it easier to train.

Although the objective of this paper is not to achieve a breakthrough in tuning performance, the performance of the proposed parameter selection combined with an explainable tuning model still slightly surpasses the current SOTA method, CDBTune, under conditions of similar convergence. In fact, under conditions utilizing a similar reinforcement learning architecture, the performance differences between the models are not significant. However, KnobTree is able to provide reasonable explanations for tuning strategies without compromising model performance, granting it a substantial advantage in practical industrial deployment.

\subsection{Ablation Experiment on Parameter Selection}

This section presents an ablation study to clarify the impact of parameter selection on model performance. We compare "KnobTree," our full model utilizing a reduced action space through parameter selection, against "KnobTree$^-$," which lacks parameter selection and uses a full action space. The concern might be that narrowing the action space through parameter selection could limit the model's potential. However, our tests in different MySQL scenarios (RO, WO, RW) demonstrate that parameter selection does not compromise the performance ceiling of KnobTree. Instead, it improves efficiency by eliminating redundant parameters that can hinder the reinforcement learning process from achieving optimal performance.

On the other hand, this paper recorded the time required for several models to complete each step of a tuning task on MySQL. 

During the database tuning process, a complete tuning task can be divided into the following four parts:
\begin{itemize}
    \item Deployment: Deploy configuration parameters to the database and test the database performance using standard performance testing tools.
    \item Metrics Collection: Collect internal database metrics as the state and compute the reward function.
    \item Model Update: Perform calculations and backpropagation in the network.
    \item Model Reasoning: Output recommended configuration parameters based on the state.
\end{itemize}
The first two parts are independent of the model used, while the last two correspond to the backward and forward propagation of the neural network, respectively. We record the statistical results in Table~\ref{table1}, which specifically presents the throughput and latency of the RO scenario; similar trends are observed in the other two scenarios.

\begin{table}[!t]
    \centering
    \caption{Performance and efficiency comparison of whether to use parameter selection}
    \resizebox{\linewidth}{!}{
    \renewcommand{\arraystretch}{1.5} 
    \begin{tabular}{ccccc}
        \hline
        Model & KnobTree & KnobTree$^-$ & CDBTune & OtterTune\\
        \hline
        Throughput (txn/s) & 17710 & 17498 & 17252 & 16896\\
        Delay (ms) & 8.16 & 8.71 & 8.89 & 9.93\\ 
        Model Update (ms) & \textbf{25.73} & 31.71 & 34.53 & 39.72\\
        Model Reasoning (ms) & \textbf{2.21} & 3.56 & 4.63 & 12.43\\
        \hline
    \end{tabular}}
    \label{table1}
\end{table}

Statistical analysis shows that KnobTree, optimized post-parameter selection, updates and infers faster than KnobTree$^-$. This efficiency stems from a reduced action space, speeding up decision-making. Compared to CDBTune, KnobTree's tree-based actor module updates more rapidly than CDBTune's neural network. Moreover, machine learning methods like OtterTune, which predict performance over multiple configurations before selecting the best, are more time-intensive.

This paper also compares the time required for online tuning in MySQL between KnobTree and other non-reinforcement learning methods, to demonstrate the advantages of reinforcement learning in database tuning. The results are shown in Table~\ref{table2}.

\begin{table}[!t]
    \centering
    \caption{Statistics of Online Tuning Time}
    \resizebox{\linewidth}{!}{
    \renewcommand{\arraystretch}{1.5} 
    \begin{tabular}{lccc}
        \hline
        Method & Steps Required & Time per Step (min) & Total Time (min) \\
        \hline
        KnobTree & 5 & 6 & 30 \\
        OtterTune & 5 & 11 & 55 \\
        MySQLTuner & 1 & 50 & 50 \\
        DBA & 1 & 120 & 120 \\
        \hline
    \end{tabular}}
    \label{table2}
\end{table}

The comparison between KnobTree and OtterTune in Table 2 demonstrates that reinforcement learning-based methods significantly outperform traditional machine learning-based methods in executing online tasks, as the latter requires finding the optimal configuration based on the prediction model for each task. MySQLTuner, a rule-based tuning script, requires further exploration by the DBA in the direction of optimization, hence the longer duration. 

In conclusion, the interpretable tuning method proposed in this paper combined with the parameter importance analysis method can quickly recommend configuration parameters when performing actual parameter adjustment tasks, helping to alleviate the pressure on the database system in production.

\subsection{Interpretability Case Study}

This subsection illustrates the explainability of KnobTree proposed in this paper through two scenarios.

\subsubsection{Scenario One} 
\begin{figure}[t]
\centering
\includegraphics[width=1\columnwidth]{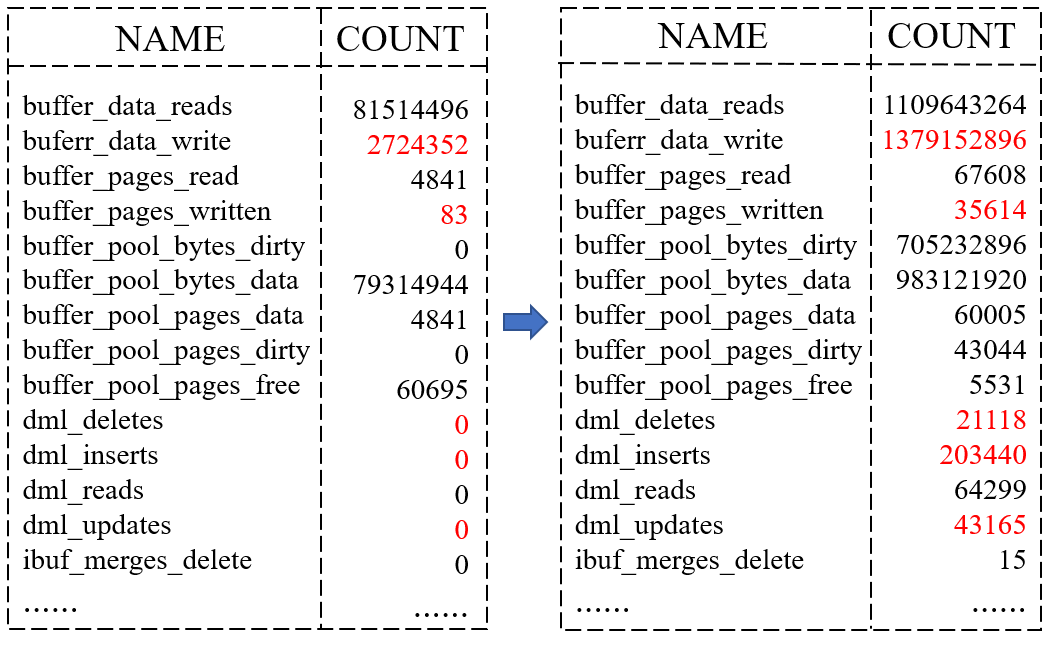} 
\caption{A database scenario change instance}
\label{fig9}
\end{figure}

\begin{figure}[t]
\centering
\includegraphics[width=0.9\columnwidth]{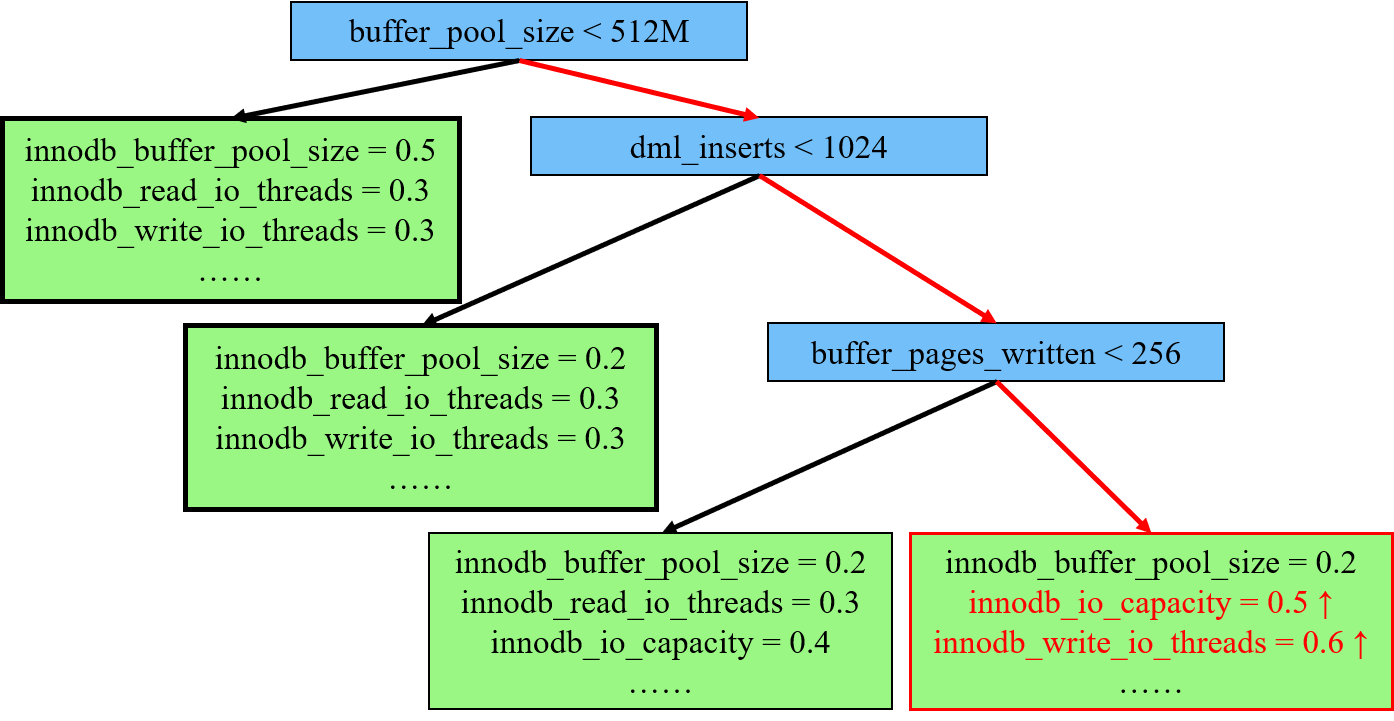} 
\caption{Policy interpretation tree in the first scenario}
\label{fig10}
\end{figure}

This scenario is used to verify that KnobTree can provide reasonable explanations for tuning strategies. We collect the performance state variables of the system over a period of time, with an example of database scene change shown in Figure~\ref{fig9}.
The variables \texttt{buffer\_data\_read}, \texttt{dml\_inserts}, and \texttt{dml\_deletes} indicate that write transactions in the system are increasing. In this scenario, the tuning strategy explanation tree generated by KnobTree in this paper is shown in Figure~\ref{fig10}. For ease of display, the tree structure has been simplified, and approximate values have been taken for the parameters in the tree.

The leaf node at the bottom right of the tree represents the tuning strategy made in this scenario. The tuning strategy can be explained by the path from the root node to this leaf node:
\begin{enumerate}
    \item The root node examines the system state variable \texttt{buffer\_pool\_size} to determine if the current buffer pool size meets the system's requirements. If the value of \texttt{buffer\_pool\_size} is less than the threshold of 512M, it implies a need to increase the buffer pool. In this case, it is found that the buffer pool is already large enough and does not need adjustment.
    \item The number of pages involved in read transactions in the system, as indicated by the system state variable \texttt{buffer\_pages\_read}, is not large, suggesting that there are not many read transactions in the system at this time.
    \item The variable \texttt{buffer\_pages\_writen} is used to assess the number of write transactions. It is observed that there are relatively more write transactions in this scenario. Particularly when there is a high volume of write transactions, incrementing the number of write threads can effectively improve the overall throughput and response rate of the database. Therefore, the main focus should be on increasing \texttt{innodb\_write\_io\_threads}, i.e., the parameter for the number of write threads.
\end{enumerate}

Through such a decision-making process, DBAs can adjust and optimize the database configuration following the intuitive guidance provided by the tree structure to achieve optimal performance.

\subsubsection{Scenario two} 

This scenario is used to illustrate the guiding significance of the strategy explanation given by KnobTree for model improvement when the model performance is not satisfactory.

\begin{figure}[t]
\centering
\includegraphics[width=0.9\columnwidth]{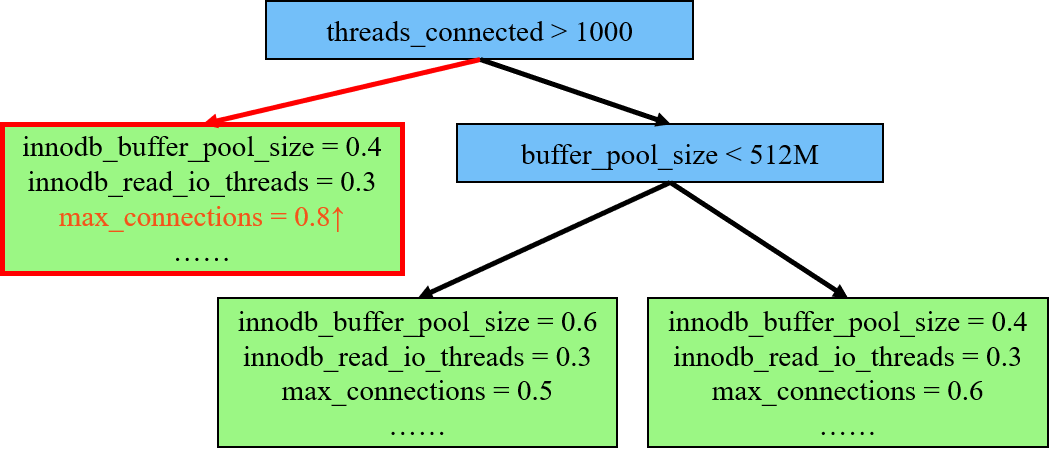} 
\caption{Policy interpretation tree in the second scenario}
\label{fig11}
\end{figure}

As illustrated in Figure~\ref{fig11}, a poor database performance resulted from the tuning strategy given by the model in the lower left corner. The tree's explanation reveals that the agent, based on the \texttt{threads\_connected} variable, tries to increase the \texttt{max\_connections} parameter. This move conflicts with the DBA's tuning experience, particularly when memory is insufficient, as each session connection competes for memory, causing performance to suffer. KnobTree's experiment highlights its ability to aid the DBA in pinpointing poor performance causes and making informed adjustments.

These scenarios demonstrate that the explainable tuning method introduced in this paper provides insightful explanations for tuning strategies, assists database administrators in enhancing performance, and serves as a valuable guide for tuning tasks.

\subsubsection{Quantitative Analysis of Tuning Strategy Interpretability}

To further validate the interpretability offered by KnobTree, this paper conducted a user survey with participants including database experts from private companies, teachers, and students engaged in database research. The participants were provided with explanations of tuning strategies extracted from trained tuning models as an aid for decision-making. These tuning models include the interpretable tuning model proposed in this paper, a neural network-based tuning model \cite{zhang2019end}, and a Gaussian process regression model \cite{zhang2018demonstration}.

\begin{figure}[t]
\centering
\includegraphics[width=0.7\columnwidth]{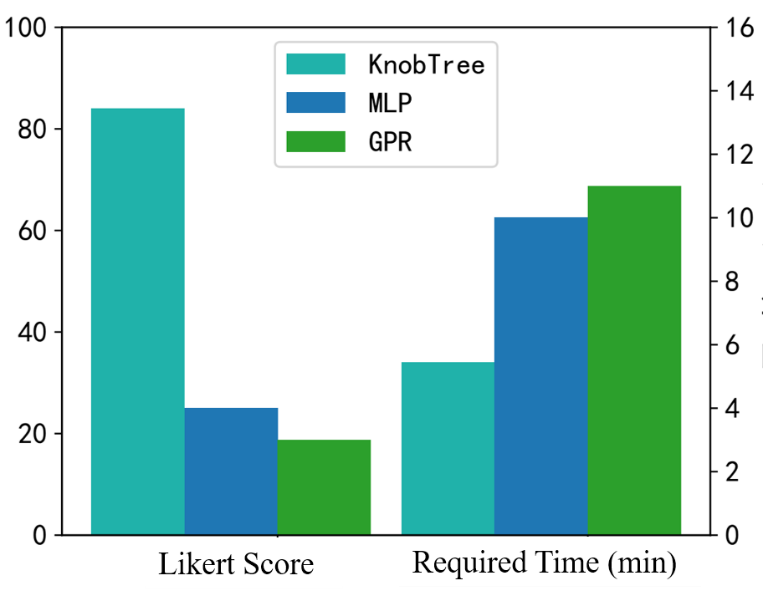} 
\caption{Interpretable user research analysis}
\label{fig12}
\end{figure}

Participants were asked to assess the interpretability and usability of the tuning strategies from the various models when used as decision-support tools. The results, as shown in Figure~\ref{fig12}, indicate that KnobTree corresponds to the method of this paper, MLP corresponds to the neural network-based tuning model, and GPR corresponds to the Gaussian process regression model. The assessment used the Likert Scale method, which is the most widely used scale in survey research, categorizing participants' evaluations into five levels, with each level corresponding to a score. Participants selected one level as their response. The figure also compiles the time participants took to make predictions when using the explanations provided by the two models as decision-support tools for tuning.

\section{Conclusion}

This paper explores the synergy between artificial intelligence and database technology, focusing on automatic database tuning through reinforcement learning. Traditional methods, while effective in learning tuning strategies in complex parameter spaces, often do not prioritize parameters based on their impact on system performance, leading to inefficient training. Our proposed model addresses this by:

\begin{enumerate}
    \item Introducing a parameter importance analysis that combines knowledge-driven approaches with Shapley values to identify and focus on the most impactful parameters. This significantly narrows the action space and speeds up the tuning process, as demonstrated by our experiments where tuning with selected parameters matches the effectiveness of comprehensive approaches in less time.
    
    \item Providing an interpretable reinforcement learning-based tuning method that not only achieves optimal tuning results but also explains the rationale behind parameter adjustments. This transparency is beneficial for both model designers and database administrators (DBAs), guiding their tuning efforts more effectively.
\end{enumerate}

Future work will consider extending the model to include parameters from the databases' operating environments and improving its adaptability to diverse workloads by incorporating workload-specific features.

\bibliographystyle{IEEEtran}
\bibliography{references}

\vfill

\end{document}